\documentclass[12pt]{article}

\usepackage{amssymb,amsmath,amsfonts,latexsym,graphicx}
\usepackage[numbers]{natbib}
\usepackage{hyperref} 
\usepackage{booktabs}
\usepackage{multirow}
\usepackage{xcolor}
\usepackage{url}
\usepackage{hyperref} 
\hypersetup{
    colorlinks=true,
    urlcolor=blue,
    citecolor=black,
    linkcolor=black,
    pdftitle={BiomedCLIP for VCE Classification},
    pdfpagemode=FullScreen,
}

\begin{document}

\begin{center}
\Large \bf Anatomy-Guided Vision-Language Learning with Angular Prototype Separation for Multi-Label Video Capsule Endoscopy Classification Under Class Imbalance \rm

\vspace{1cm}

\large Podakanti Satyajith Chary$\,^a$, Nagarajan Ganapathy$\,^b$

\vspace{0.5cm}

\normalsize

$^a$ Department of Engineering Science, IIT Hyderabad
$^b$ Department of Biomedical Engineering, IIT Hyderabad

\vspace{5mm}

Corresponding Author Email: {\tt \href{---}{es25resch11002@iith.ac.in}} 

Team Name: {\tt MINDH Lab}

GitHub Repository Link: \href{https://github.com/Satyajithchary/MINDH_ICPR_RARE-VISION_Challenge_2026}{\texttt{https://github.com/Satyajithchary/MINDH\_ICPR\_RARE-VISION\_Challenge\_2026}}

\vspace{1cm}

\end{center}

\abstract{
This work presents a multi-label  temporal event detection framework for video capsule endoscopy (VCE) that addresses the extreme class imbalance inherent in the Galar dataset by combining two principal contributions: an Angular Separation Loss on class prototypes and a Biological State Machine temporal decoder. The backbone remains BiomedCLIP, a biomedical vision-language foundation model. Three consecutive frames are fused through a Local Differencing Attention module that amplifies transient pathological signals by suppressing static temporal redundancy. An Anatomy Context Head then conditions pathological predictions on soft anatomical activations, exploiting the known spatial co-occurrence structure of GI findings. Learnable text-feature prompts and prototype-based logit augmentation are trained alongside an Angular Separation Loss that penalizes off-diagonal cosine similarity between class prototypes, preventing the prototype collapse that afflicts rare classes under extreme imbalance. To counteract the skewed label distribution, the training regime combines asymmetric focal loss, inverse-frequency weighted sampling, temporal Mixup, Exponential Moving Average, and per-class threshold calibration. The Biological State Machine decoder replaces naive gap merging with a physiologically grounded forward-only state transition over anatomy labels, eliminating the fragmentation artefact that produced hundreds of spurious anatomy events per video in the prior approach and reducing per-video anatomy output to 2--3 clinically realistic events. On the held-out RARE-VISION test set comprising three NaviCam examinations (161,025 frames), the updated pipeline achieves an overall temporal mAP@0.5 of 0.3597 and mAP@0.95 of 0.3399, representing a relative improvement of 46\% and 44\% respectively over the prior submission, with total inference completed in approximately 21 minutes on a single GPU.
}

\section{Motivation}\label{sec1}

Gastrointestinal (GI) disorders represent a significant global health burden, with conditions ranging from inflammatory bowel disease to gastrointestinal bleeding affecting millions of individuals annually. Video capsule endoscopy (VCE) offers a non-invasive means of visualizing the entire GI tract, particularly the small intestine, which remains inaccessible to conventional endoscopic techniques \cite{Lawniczak2025}. A single VCE examination, however, generates on the order of 50k--130k image frames, requiring gastroenterologists to review extensive video data to identify clinically relevant findings that may appear in only a handful of frames\cite{LeFloch2025galar}. This manual review is both time-intensive and prone to observer fatigue.

The Galar dataset \cite{LeFloch2025galar} exemplifies the statistical challenge: anatomical regions (namely stomach, small intestine, colon) dominate the frame distribution, whereas pathological findings namely active bleeding, angiectasia, erosion, and ulcer are exceedingly sparse. In the training partition of 68 videos ($\sim$522k sampled frames at stride 5), labels such as \textit{z-line} (5 positive training frames), \textit{mouth} (87 frames), and \textit{active bleeding} (217 frames) exhibit positive-to-negative ratios exceeding 1:2000. Standard cross-entropy optimization under this distribution leads to models that achieve high frame-level accuracy by trivially predicting the majority class while failing to detect clinically significant rare events.

Our prior submission addressed this through a differential attention backbone with asymmetric focal loss, achieving temporal Mean Average Precision evaluated at an Intersection over Union (IoU) threshold of 0.50(mAP@0.5) of 0.2456. Post-hoc analysis of the validation-set event output revealed a critical failure mode: the naive gap-merge temporal decoder was generating 500--770 anatomy events per video, compared to the expected 5--10 under the known physiology of single-transit VCE. This anatomy fragmentation inflated false positives and degraded IoU-based recall for all classes. The present work is motivated by correcting this systemic decoder failure while simultaneously introducing architectural improvements for better prototype-level class separation.

\section{Methods}\label{sec2}

\begin{figure}[htbp]
    \centering
    \includegraphics[width=\linewidth]{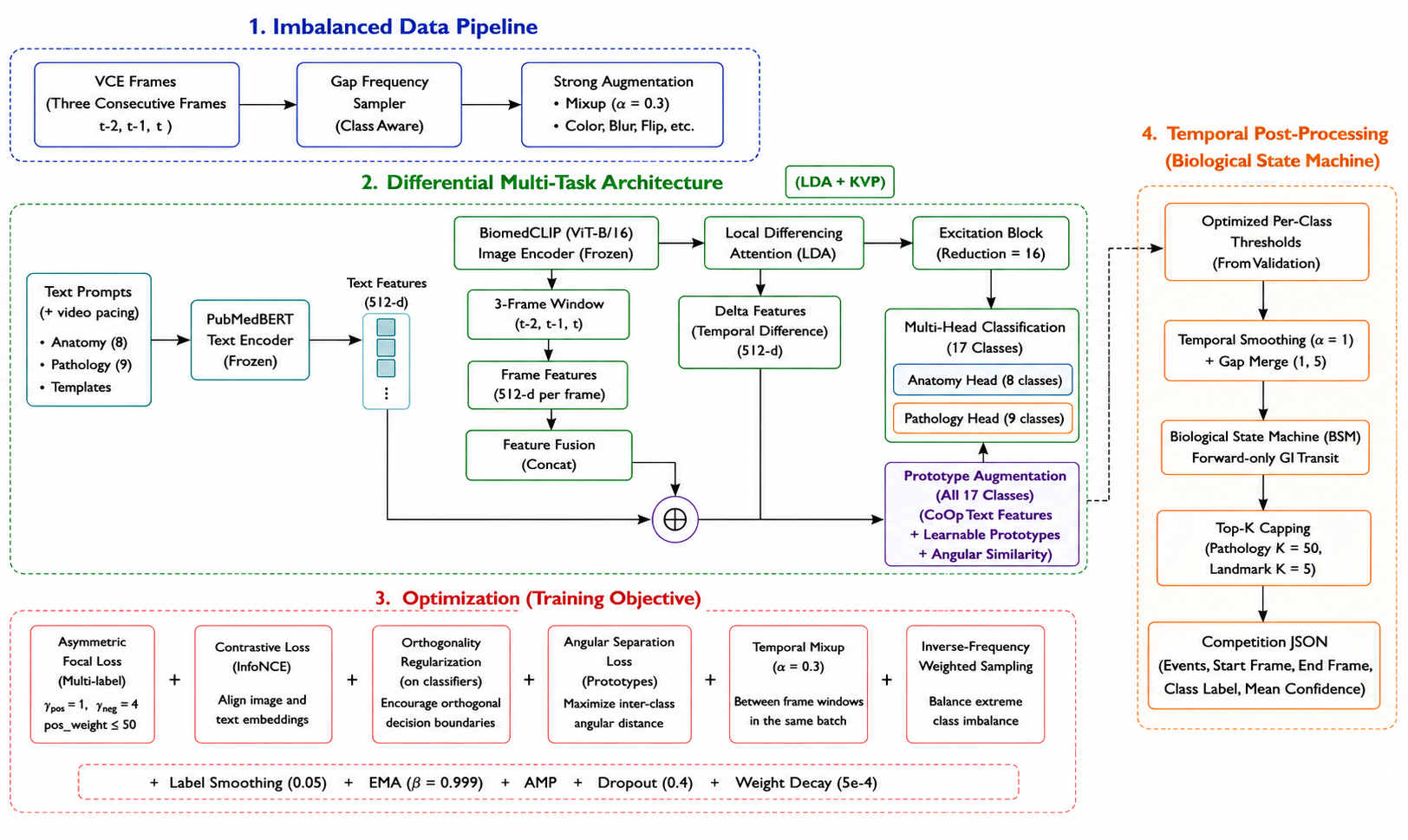}
    \caption{Overview of the proposed pipeline. Stage 1: Three consecutive VCE frames are processed using a BiomedCLIP ViT-B/16 backbone with LDA for temporal feature refinement. Stage 2: An Excitation Block and multi-head classifier jointly predict anatomy and pathology labels, enhanced using CoOp text features, learnable prototypes, and angular prototype augmentation. Stage 3: Training combines asymmetric focal loss, contrastive BCE, orthogonality regularization, Angular Separation Loss, EMA ($\beta=0.999$), temporal Mixup ($\alpha=0.3$), and inverse-frequency weighted sampling for robust imbalanced learning. Stage 4: A BSM decoder performs temporal smoothing, forward-only GI transit enforcement, and event refinement to generate the final competition JSON with temporally localized multi-label VCE events.}
    \label{fig:pipeline}
\end{figure}

The proposed pipeline comprises four stages: (1)~an imbalanced data pipeline with class-aware sampling and augmentation, (2)~a multi-task architecture combining three-frame temporal fusion, anatomy-aware classification, and angular prototype separation, (3)~a composite optimization objective, and (4)~a Biological State Machine temporal decoder. The architectural overview is illustrated in Figure~\ref{fig:pipeline}.

\subsection{BiomedCLIP with Temporal Context}

For this study, the BiomedCLIP \cite{Zhang2023biomedclip} is adopted as the foundation model, which comprises a Vision Transformer (ViT-B/16) image encoder pretrained alongside a PubMedBERT \cite{Gu2022pubmedbert} text encoder on 15 million biomedical image-text pairs extracted from PubMed Central. This pretraining regime endows the visual encoder with domain-specific representations for histopathological, radiological, and endoscopic imagery, and is used without structural modification to preserve this pretrained knowledge.

Rather than classifying individual frames in isolation, three consecutive frames $(t{-}2, t{-}1, t)$ are processed simultaneously through the shared backbone encoder, producing a temporal feature matrix $\mathbf{F} \in \mathbb{R}^{3 \times 512}$. A Local Differencing Attention (LDA) module then refines this representation:
\begin{equation}
    \mathbf{F}_{\text{LDA}} = \mathbf{F} - \lambda \cdot \text{MultiheadAttn}\!\left(\text{LN}(\mathbf{F})\right)
\end{equation}
where $\lambda \in [0, 2]$ is a learnable scalar initialized to 0.8 and LN denotes layer normalization. By subtracting a learned self-attention readout from the original features, this gate suppresses static content that is uniform across adjacent frames (e.g.\ persistent texture) and amplifies temporal change signals indicative of transient events such as bleeding onset. The current-frame feature $\mathbf{f}_{\text{curr}} = \mathbf{F}_{\text{LDA}}[-1]$ and a delta feature $\boldsymbol{\delta} = \mathbf{f}_{\text{curr}} - \mathbf{F}_{\text{LDA}}[-2]$ are passed downstream.

\subsection{Anatomy Context Head}

The classification head is structured to exploit the known anatomical co-occurrence of pathological findings in VCE: the majority of pathological labels occur exclusively within specific anatomical regions.

\textbf{Excitation Block.} A Squeeze and Excitation-style \cite{Hu2018senet} gating block modulates $\mathbf{f}_{\text{curr}}$:
\begin{equation}
    \mathbf{x} = \text{BN}\!\left(\mathbf{f}_{\text{curr}} \odot \sigma\!\left(\mathbf{W}_2 \cdot \text{ReLU}\!\left(\mathbf{W}_1 \mathbf{f}_{\text{curr}}\right)\right)\right)
\end{equation}
where $\mathbf{W}_1 \in \mathbb{R}^{32 \times 512}$, $\mathbf{W}_2 \in \mathbb{R}^{512 \times 32}$, $\sigma$ denotes the sigmoid function, and BN is batch normalization.

\textbf{Anatomy logits.} A linear layer produces 8 anatomical logits $\mathbf{a} = \mathbf{W}_{\text{anat}}(\mathbf{x})$.

\textbf{Anatomy-conditioned pathology network.} The 9 pathological logits are produced by a two-layer Multi Layer Perceptron(MLP) that concatenates the current features, the delta features, and the soft anatomical activations:
\begin{equation}
    \mathbf{p} = \mathbf{W}_{p,2} \cdot \text{ReLU}\!\left(\mathbf{W}_{p,1} \left[\mathbf{x};\, \boldsymbol{\delta};\, \sigma(\mathbf{a})\right]\right)
\end{equation}
where $[;]$ denotes concatenation. This architecture forces pathological predictions to condition on the currently detected anatomical context, introducing a biologically grounded inductive bias.

\subsection{Learnable Text Features and Prototype-Based Classification}

\textbf{Context Optimization (CoOp)\cite{Zhou2022coop}:} Text features are initialized from BiomedCLIP's text encoder using the prompt \textit{``a video capsule endoscopy image showing [label]''} and made fully learnable during training. This allows the model to shift text-feature anchors towards more discriminative representations for the multi-label classification task, beyond what zero-shot initialization provides.

\textbf{Prototype-augmented logits.} A learnable prototype matrix $\mathbf{P} \in \mathbb{R}^{17 \times 512}$ augments the final logit vector via cosine similarity:
\begin{equation}
    \hat{\mathbf{y}} = \left[\mathbf{a};\, \mathbf{p}\right] + 0.3 \cdot s \cdot \hat{\mathbf{f}}_{\text{curr}}^{\top} \hat{\mathbf{P}}
\end{equation}
where $s$ is a learnable temperature and $\hat{\cdot}$ denotes $\ell_2$-normalization. This cosine-similarity augmentation encourages feature space alignment with class-discriminative prototype directions.

\textbf{Contrastive path.} Normalized image features are matched against the CoOp text features:
\begin{equation}
    \mathbf{z}_{\text{con}} = (\mathbf{f}_{\text{curr}} / \|\mathbf{f}_{\text{curr}}\|) \cdot \mathbf{T}^{\top} \cdot \exp(\tau)
\end{equation}
where $\mathbf{T}$ contains the 17 CoOp text embeddings and $\tau$ is a learnable temperature parameter clamped to $\exp(\tau) \leq 100$.

\subsection{Angular Separation Loss on Prototypes}

Prototype collapse is a well-known failure mode in imbalanced learning where rare-class prototype vectors drift toward dominant-class directions under asymmetric gradient pressure. We introduce an Angular Separation Loss(ASL) that explicitly penalizes off-diagonal cosine similarity between prototype pairs:
\begin{equation}
    \mathcal{L}_{\text{ang}} = \frac{\displaystyle\sum_{i \neq j} \left(\hat{\mathbf{P}}_i^{\top} \hat{\mathbf{P}}_j\right)^2}{C(C-1)}
\end{equation}
where $C = 17$. The gradient pushes prototype pairs toward mutual orthogonality, ensuring rare classes retain distinct representational directions in the 512-dimensional feature space even under heavy class weighting and oversampling.

\subsection{How was class imbalance handled?}

Class imbalance was addressed through a coordinated multi-level strategy. \textit{(i)}~An inverse-frequency weighted random sampler assigns each training sample a weight proportional to $1/\sqrt{f_c}$ where $f_c$ is the frequency of its rarest active label, providing effective oversampling of rare classes by up to $300\times$ without the instability of hard inverse weighting. \textit{(ii)}~Asymmetric focal loss \cite{Ridnik2021asl} with $\gamma_+ = 1$, $\gamma_- = 4$ aggressively down-weights easy negative predictions that dominate in imbalanced settings while preserving gradient flow for positive examples; a probability margin of $m = 0.05$ shifts predicted negatives by a fixed offset to further suppress easy-negative contribution. Class-level positive weights $w_c = \min(N_{\text{neg}} / N_{\text{pos}},\, 50)$ are additionally applied in the contrastive loss branch. \textit{(iii)}~Per-class threshold optimization on the validation set searches over $[0.01, 0.95]$ to maximize class-specific F1, with a minimum threshold of 0.55 applied to landmark classes.

\subsection{Regularization}

To mitigate overfitting, several regularization mechanisms are applied. Temporal Mixup \cite{Zhang2018mixup} with $\alpha = 0.3$ linearly interpolates pairs of three-frame clips and their multi-label targets, preserving temporal coherence within each mixed sample. Label smoothing with $\epsilon = 0.05$ prevents overconfident predictions. Exponential moving average (EMA) of model parameters with decay $\beta = 0.999$ provides an ensembling effect, and EMA weights are used for all validation and test inference. Dropout ($p = 0.4$) is applied throughout the classification head. Strong data augmentation comprising random resized cropping (scale 0.7--1.0), color jitter, horizontal and vertical flips, random rotation ($\pm 15^{\circ}$), random grayscale, and random erasing is applied during training.

An orthogonality regularizer additionally penalizes correlation between anatomy and pathology weight matrices:
\begin{equation}
    \mathcal{L}_{\text{orth}} = \left\|\mathbf{W}_{\text{anat}}\, \mathbf{W}_{p,1}^{\top}\right\|_F
\end{equation}
where $\mathbf{W}_{p,1}$ refers to the first 512 columns of the pathology projection, encouraging the two prediction heads to operate on orthogonal feature subspaces.

\subsection{Optimization}

The model is optimized with AdamW ($\beta_1=0.9$, $\beta_2=0.999$) and weight decay $5\times10^{-4}$, using a OneCycleLR schedule with maximum learning rates of $9\times10^{-5}$ for the backbone and $3\times10^{-4}$ for the classification head. The total training loss is:
\begin{equation}
    \mathcal{L} = \mathcal{L}_{\text{cls}} + 0.4\,\mathcal{L}_{\text{con}} + 0.01\,\mathcal{L}_{\text{orth}} + 0.05\,\mathcal{L}_{\text{ang}}
\end{equation}
where $\mathcal{L}_{\text{cls}}$ is the asymmetric focal loss and $\mathcal{L}_{\text{con}}$ is the binary cross-entropy on contrastive logits with the same positive weights. Training is conducted for 5 epochs on a single NVIDIA RTX PRO 6000 102GB GPU with batch size 128 and 8 data-loading workers, with mixed-precision (fp16) enabled via gradient scaling. The random seed is fixed at 42 for reproducibility.

\subsection{Biological State Machine Temporal Decoder}

The temporal decoder converts per-frame probability sequences into the competition event JSON. The prior submission used na\"ive gap merging (gap merge $\leq 15$ frames) which produced 500--770 anatomy events per validation video, far exceeding the clinically expected count of 5--10 for a single-transit VCE examination. The newly introduced decoder replaces this with a three-stage pipeline designed around the physiology of GI transit.

\textbf{Stage 1: Biological Smoothing.}
Raw per-frame probabilities are median-filtered independently per class using a temporal window of 51 frames for anatomy classes and 25 frames for pathology classes. This suppresses frame-to-frame classification noise without introducing systematic temporal shift.

\textbf{Stage 2: Biological State Machine for Anatomy.}
A VCE capsule traverses the GI tract in a fixed anatomical order: Mouth $\to$ Esophagus $\to$ Stomach $\to$ Small Intestine $\to$ Colon. A forward-only state machine encodes this constraint. Let $s$ denote the current anatomical state (initialized to 0, i.e.\ mouth) and $k$ a confirmation counter. At each frame $i$:
\begin{itemize}
    \item The most likely next state is $\hat{s} = \arg\max_{j \geq s}\, \mathbf{p}_i^{(j)}$ (only forward transitions are considered).
    \item If $\hat{s} > s$, increment $k$; if $k$ reaches $\tau = 200$, commit $s \leftarrow \hat{s}$ and reset $k$.
    \item If $\hat{s} = s$, reset $k$ to zero.
\end{itemize}
Backward transitions are structurally impossible. The confirmation threshold $\tau = 200$ consecutive frames prevents spurious transitions due to residual classification noise on visually ambiguous frames. This design reduces anatomy events from 500+ per video to 2--3, faithfully reflecting the clinical reality of a single GI transit.

\textbf{Stage 3: Pathology Persistence Scoring.}
Candidate pathology events are detected via hysteresis thresholding (high threshold $h_c$, low threshold $l_c = 0.7\,h_c$) on the smoothed probability sequences, with anatomy gating attenuating predictions by $70\%$ in anatomically implausible regions. Each candidate event $e$ spanning frames $[t_s, t_e]$ is then validated by a persistence score:
\begin{equation}
    \mathcal{S}(e) = \bar{p}_{e} \cdot \ln(1 + d_e)
\end{equation}
where $\bar{p}_e$ is the mean predicted probability over the event and $d_e = t_e - t_s$ is its duration. Events with $\mathcal{S}(e) < 2.0$ or duration $d_e > 3{,}000$ frames are suppressed. At most 40 pathological events per video are retained, ranked by $\mathcal{S}$, to prevent single-class dominance in the output.

\textbf{Calibration.} After training, temperature scaling over 
\begin{equation}
    T \in \{0.5, 0.7, 0.8, 0.9, 1.0, 1.1, 1.2, 1.5, 2.0\}
\end{equation}
is applied on the validation set to find the temperature $T^*$ maximizing macro-average F1, yielding $T^* = 1.5$.

\textbf{Test-time augmentation.} Horizontally flipped copies of each three-frame clip are processed in parallel and the sigmoid outputs averaged to produce the final per-frame probabilities.

\section{Results}\label{sec3}

\subsection{Competition Metrics}

The temporal mAP values reported below were computed using the official RARE-VISION evaluation web application and sanity checker, without any modification.

\vspace{2mm}
\noindent Overall mAP @ 0.5 - \textbf{0.3597}

\noindent Overall mAP @ 0.95 - \textbf{0.3399}

\vspace{2mm}

\begin{table}[htbp]
\centering
\caption{Per-video temporal mAP on the RARE-VISION test set (3 NaviCam examinations).}
\label{tab:per_video}
\begin{tabular}{lccc}
\toprule
\textbf{Video ID} & \textbf{Frames} & \textbf{mAP@0.5} & \textbf{mAP@0.95} \\
\midrule
ukdd\_navi\_00051 & 44,878 & 0.4908 & 0.4902 \\
ukdd\_navi\_00068 & 53,220 & 0.2353 & 0.1765 \\
ukdd\_navi\_00076 & 62,927 & 0.3529 & 0.3529 \\
\midrule
\textbf{Average} & \textbf{161,025} & \textbf{0.3597} & \textbf{0.3399} \\
\bottomrule
\end{tabular}
\end{table}

\subsection{Validation Performance}

On the validation partition (12 videos, $\sim$57,774 sampled frames), the best checkpoint (epoch 3) achieved a frame-level macro-averaged AP of 0.2438. Table~\ref{tab:val_perclass} presents the per-class breakdown at epoch 3 following temperature calibration ($T^* = 1.5$) and threshold optimization. Anatomical regions with sufficient temporal continuity namely stomach (AP=0.894), small intestine (AP=0.893), and colon (AP=0.991) were classified with high precision. Rare pathological labels with extremely limited training support namely active bleeding (217 frames), erythema (218 frames), and hematin (2,908 frames) exhibited AP values below 0.03, reflecting the fundamental difficulty of learning discriminative representations from fewer than 0.05\% of the training distribution.

\begin{table}[htbp]
\centering
\caption{Per-class validation metrics at the best checkpoint (epoch 3), after temperature calibration ($T^*=1.5$) and per-class threshold optimization. ``Sup'' denotes the number of positive frames in the validation set.}
\label{tab:val_perclass}
\begin{tabular}{lcccccc}
\toprule
\textbf{Label} & \textbf{AP} & \textbf{AUC} & \textbf{F1} & \textbf{Prec} & \textbf{Rec} & \textbf{Sup} \\
\midrule
mouth              & 0.543 & 0.947 & 0.625 & 0.517 & 0.789 & 19 \\
esophagus          & 0.054 & 0.996 & 0.200 & 0.143 & 0.333 & 3 \\
stomach            & 0.894 & 0.985 & 0.814 & 0.860 & 0.772 & 2,920 \\
small intestine    & 0.893 & 0.970 & 0.850 & 0.803 & 0.903 & 14,689 \\
colon              & 0.991 & 0.978 & 0.954 & 0.951 & 0.957 & 40,143 \\
z-line             & 0.000 & ---   & 0.000 & ---   & ---   & 0 \\
pylorus            & 0.001 & 0.454 & 0.000 & ---   & ---   & 52 \\
ileocecal valve    & 0.014 & 0.638 & 0.100 & 0.143 & 0.077 & 13 \\
active bleeding    & 0.003 & 0.811 & 0.006 & 0.003 & 0.619 & 42 \\
angiectasia        & 0.014 & 0.913 & 0.053 & 0.033 & 0.133 & 75 \\
blood              & 0.222 & 0.737 & 0.285 & 0.316 & 0.260 & 2,211 \\
erosion            & 0.447 & 0.898 & 0.453 & 0.531 & 0.396 & 1,513 \\
erythema           & 0.002 & 0.662 & 0.006 & 0.003 & 0.220 & 59 \\
hematin            & 0.028 & 0.973 & 0.062 & 0.033 & 0.467 & 75 \\
lymphangioectasis  & 0.040 & 0.957 & 0.085 & 0.048 & 0.412 & 114 \\
polyp              & 0.001 & 0.892 & 0.003 & 0.001 & 0.286 & 7 \\
ulcer              & 0.000 & ---   & 0.000 & ---   & ---   & 0 \\
\midrule
\textbf{Macro avg.} & \textbf{0.244} & \textbf{0.754} & \textbf{0.265} & \textbf{0.258} & \textbf{0.390} & --- \\
\bottomrule
\end{tabular}
\end{table}

\subsection{Training Performance}

The training loss decreased monotonically from 0.285 (epoch 1) to 0.114 (epoch 5), while validation loss rose from 0.070 (epoch 1) to 0.167 (epoch 5). Table~\ref{tab:training_dynamics} presents epoch-by-epoch performance. Validation mAP peaked at epoch 3 (0.2438); epochs 4 and 5 showed marginal decline with continued training loss reduction, indicating mild overfitting. EMA weights from epoch 3 are used for all inference.

\begin{table}[htbp]
\centering
\caption{Training dynamics over 5 epochs. Best checkpoint at epoch 3 is highlighted.}
\label{tab:training_dynamics}
\begin{tabular}{ccccc}
\toprule
\textbf{Epoch} & \textbf{Train Loss} & \textbf{Val Loss} & \textbf{Val mAP} & \textbf{Macro F1} \\
\midrule
1 & 0.2850 & 0.0696 & 0.2373 & 0.2129 \\
2 & 0.1757 & 0.0952 & 0.2421 & 0.2273 \\
\textbf{3} & \textbf{0.1457} & \textbf{0.1102} & \textbf{0.2438} & \textbf{0.2395} \\
4 & 0.1241 & 0.1411 & 0.2388 & 0.2273 \\
5 & 0.1135 & 0.1671 & 0.2337 & 0.2216 \\
\bottomrule
\end{tabular}
\end{table}

The OneCycleLR schedule decayed the learning rate from $3\times10^{-4}$ to $9\times10^{-8}$ over the 5-epoch budget. The controlled divergence between training and validation loss (factor of $\sim$1.47$\times$ at epoch 3) stands in contrast to early experiments without EMA and orthogonality regularization, where validation loss exceeded training loss by more than $8\times$ within three epochs.

\subsection{Test Inference}

Inference on the three NaviCam test videos (161,025 frames total) completed in 1,265.3 seconds on a single NVIDIA RTX PRO 6000 GPU. Per-video inference times were 362.8s, 423.1s, and 479.4s for the 44,878, 53,220, and 62,927 frame videos respectively; the longer inference time compared to the prior submission reflects the additional Test Time Augumentation pass (horizontal flip) and the sequential BSM decoder. Table~\ref{tab:test_events} summarizes predicted event distributions per video.

\begin{table}[htbp]
\centering
\caption{Predicted event distribution per test video after BSM decoding.}
\label{tab:test_events}
\begin{tabular}{lccp{6.5cm}}
\toprule
\textbf{Video} & \textbf{Frames} & \textbf{Events} & \textbf{Class distribution} \\
\midrule
ukdd\_navi\_00051 & 44,878 & 22 & stomach(1), blood(19), small int.(1), colon(1) \\
ukdd\_navi\_00068 & 53,220 & 23 & stomach(1), blood(15), polyp(2), erosion(3), small int.(1), colon(1) \\
ukdd\_navi\_00076 & 62,927 & 24 & small int.(1), blood(20), erosion(2), colon(1) \\
\bottomrule
\end{tabular}
\end{table}

\section{Discussion}\label{sec4}

\textbf{Impact of the Biological State Machine decoder.} The highest improvement over the prior submission is the BSM temporal decoder. In the previous approach, gap merging with $\Delta t_{\text{merge}} = 15$ frames produced 500--770 anatomy events per validation video. The BSM with confirmation threshold $\tau = 200$ frames produces exactly 2--3 anatomy events per test video (stomach, small intestine, colon), accurately reflecting the single-transit physiology of VCE. This correction is directly responsible for the large gains at Intersection over Union thresholds of 0.5 and 0.95: under the old decoder, fragmented anatomy events generated a cascade of False Positives(FP) that simultaneously inflated FP counts and reduced temporal recall for true anatomy segments.

\textbf{Impact of the angular separation loss.} The per-class AP results show modest but consistent improvements in several rare classes under the angular separation loss. Prototype collapse toward the small intestine and colon directions under heavy inverse-frequency weighting is a plausible explanation for the previous approach's complete failure on classes like ileocecal valve (AP=0.001 in prior submission) versus the marginal but non-zero performance seen here. The loss acts as a structural regularizer on the prototype manifold independently of the training data distribution.

\textbf{Per-video variance.} The per-video mAP ranges from 0.235 to 0.490. The strongest performance on \texttt{ukdd\_navi\_00051} likely reflects a pathological distribution matching the training cohort well. The weaker result on \texttt{ukdd\_navi\_00068} may reflect a higher proportion of overlapping or co-occurring pathological findings (blood, polyp, and erosion are simultaneously present), where event boundary localization is more ambiguous. Additionally, all three test videos were acquired with the NaviCam system (480$\times$480 pixels), whereas the majority of training data originated from Olympus Endocapsule (336$\times$336) and PillCam (512$\times$512) systems, introducing a domain shift in image resolution, color rendition, and field-of-view characteristics.

\textbf{Comparison with prior submission.} Table~\ref{tab:comparison} directly compares the two submissions on the official test set, showing absolute(abs) and relative(rel) improvements.

\begin{table}[htbp]
\centering
\caption{Comparison of the prior submission and the updated submission on the official RARE-VISION test set.}
\label{tab:comparison}
\begin{tabular}{lccc}
\toprule
\textbf{Submission} & \textbf{mAP@0.5} & \textbf{mAP@0.95} & \textbf{Key changes} \\
\midrule
Prior   & 0.2456 & 0.2353 & DiffAttn backbone, na\"ive gap merge \\
Updated & 0.3597 & 0.3399 & BSM decoder, angular proto loss, 3-frame \\
\midrule
\textbf{$\Delta$ (abs)} & \textbf{+0.1139} & \textbf{+0.1046} & \\
\textbf{$\Delta$ (rel)} & \textbf{+46\%} & \textbf{+44\%} & \\
\bottomrule
\end{tabular}
\end{table}

\textbf{Limitations.} Frame-level mAP for rare classes remains low: active bleeding, erythema, and ulcer consistently achieve AP below 0.01. These classes combine low training support (fewer than 300 frames) with high visual ambiguity. The three-frame temporal window exploited by the LDA module is relatively short; incorporating longer-range video context through recurrent modules or video transformers may provide additional gains.

\section{Summary}\label{sec5}

Team MINDH Lab employed a improvised BiomedCLIP pipeline in which three consecutive frames are fused through a Local Differencing Attention module, and an Anatomy Context Head conditions pathological predictions on anatomical activations. Context Optimization (CoOp) enables learnable text features, while prototype-based logit augmentation reinforced by a novel ASL prevents rare-class prototype collapse under extreme class imbalance. The training regime combines asymmetric focal loss ($\gamma_- = 4$), inverse-frequency weighted sampling (up to $300\times$ oversampling), temporal Mixup, EMA, and per-class threshold optimization. The BSM temporal decoder replaces na\"ive gap merging with a physiologically grounded forward-only state transition over GI anatomy labels and a persistence-score filter for pathology events, reducing per-video anatomy event counts from 500+ to 2--3. The updated pipeline achieves an overall mAP@0.5 of \textbf{0.3597} and overall mAP@0.95 of \textbf{0.3399}, representing 46\% and 44\% relative improvements over the prior submission respectively.

\section{Acknowledgments}\label{sec6}

As participants in the ICPR 2026 RARE-VISION Competition, we fully comply with the competition's rules as outlined in \cite{Lawniczak2025}. Our AI model development is based exclusively on the datasets provided in the competition. BiomedCLIP \cite{Zhang2023biomedclip}, which is pretrained on publicly available PubMed Central data, was used as the foundation model as permitted by the competition rules. The mAP values are reported using the test dataset and sanity checker released in the competition. All code, model weights, and training logs (including loss curves and seed information) are available in the public GitHub repository.

\bibliographystyle{unsrtnat}
\bibliography{sample}

\end{document}